\documentclass[runningheads,a4paper]{llncs}

\usepackage{amssymb}
\usepackage{graphicx}
\usepackage{gensymb}
\usepackage{float}
\usepackage{dsfont}
\usepackage{url}
\usepackage{amsmath}

\urldef{\mailsa}\path|{madcanda, edizquie}@indiana.edu|
\newcommand{\keywords}[1]{\par\addvspace\baselineskip
\noindent\keywordname\enspace\ignorespaces#1}

\begin{document}

\mainmatter  % start of an individual contribution

\title{Information Bottleneck in Control Tasks with Recurrent Spiking Neural Networks}
%\title{Information Bottleneck in Recurrent Spiking Neural Networks for Non-Markovian Pole Balancing}

\titlerunning{Information Bottleneck in Spiking RNNs}

\author{Madhavun Candadai Vasu \and Eduardo J. Izquierdo}
\authorrunning{Information Bottleneck in Spiking RNNs: Candadai Vasu et. al.}

\institute{Cognitive Science, School of Informatics and Computing\\
Indiana University, Bloomington, U.S.A.\\
\mailsa\\
%\url{http://www.springer.com/lncs}
}


\toctitle{Info. Bottleneck in control tasks}
\tocauthor{Candadai Vasu et. al.}
\maketitle

\begin{abstract}
%Understanding information flow in neural circuits are
The nervous system encodes continuous information from the environment in the form of discrete spikes, and then decodes these to produce smooth motor actions. Understanding how spikes integrate, represent, and process information to produce behavior is one of the greatest challenges in neuroscience. Information theory has the potential to help us address this challenge. Informational analyses of deep and feed-forward artificial neural networks solving static input-output tasks, have led to the proposal of the \emph{Information Bottleneck} principle, which states that deeper layers encode more relevant yet minimal information about the inputs. Such an analyses on networks that are recurrent, spiking, and perform control tasks is relatively unexplored. Here, we present results from a Mutual Information analysis of a recurrent spiking neural network that was evolved to perform the classic pole-balancing task. Our results show that these networks deviate from the \emph{Information Bottleneck} principle prescribed for feed-forward networks.

\end{abstract}

\section{Introduction}

% 1. Deep Learning
Deep Learning systems have surpassed other algorithms and even humans at several tasks~\cite{Bengio:2013,Krizhevsky:2012,Mnih:2013,Silver:2016}. While their applications continue to grow, deep learning systems are still considered black-box optimization methods. One of the most vital features behind their success is their ability to extract relevant yet minimal information as it progresses into deeper and deeper layers \cite{Tishby:2000}. This is an extension of \emph{Rate Coding Theory}~\cite{Berger:1971} presented as the \emph{Information Bottleneck} principle~\cite{Tishby:2000,Tishby:2015}. The information bottleneck principle has been primarily focused on systems that are (a)~feedforward, and (b)~in an open-loop, decoupled from their environment.
%This is of technical importance because in control tasks the layer of the network that exercises control typically impacts the input variables. As a result, they would have high information content on those. Hence, the \emph{Information Bottleneck} principle for control tasks can be stated as an information filtering process to extract minimal relevant information followed by an expansion of information that happens at the control decision making.

% 2. Neuroscience
Neuroscientists, on the other hand, have long been studying the principles behind encoding and representation of environmental information in neural activity using principles of information theory~\cite{Borst:1999} and rate distortion theory~\cite{Simoncelli:2001}. Continuous variables from the environment are encoded as discrete spikes in the brain, which are then decoded to produce smooth continuous movement. Due to experimental limitations, an informational analysis of a closed-loop brain-body-environment behaviour system is not yet feasible.

% 3. Our approach: Task
We take a radically different approach to understanding information flow in a behaviorally-functional agent. We artificially evolve embodied agents controlled by recurrent spiking neural networks to perform a task. For this paper, we focus on a non-Markovian version of the classical pole balancing task. Pole balancing has been explored quite extensively as a benchmark for control using neural networks~\cite{Pasemann:1997,Onat:1998}. With continuous states and actions, this  task serves as an ideal setting to study the transformation of the continuous input information into spikes and then back to continuous action. While the typical task is Markovian, and thus too trivial for an informational analysis, it can easily be transformed into a non-Markovian task by making the available information to the agent limited. Our approach to pole balancing incorporates an agent-centric ``visual'' interface to the pole. Therefore, information that is typically available, like the pole's angle and angular velocity, the agent's position and velocity, are not fed directly to the network. Ultimately, the minimal nature of the task makes it tractable for an investigation of a recurrent network in a closed-loop task.

% 4. Approach: Optimization
The parameters of the recurrent spiking neural network that balances the pole were optimized using an evolutionary algorithm. Evolving neural networks as opposed to hand-designing them allows maximum flexibility for exploring the parameter space. While evolutionary algorithms have been very commonly used in several fields~\cite{Jong:2006}, recently, they have been proven to be efficient for optimizing deep neural networks as well~\cite{Miikkulainen:2017,Salimans:2017}. Moreover, due to the stochastic nature of the optimization, running the algorithm several times provides us with an ensemble of solutions that solve the problem. This allows the analysis of not just one solution but several to evaluate consistency of results.

% 5. Organization
The paper is organized as follows. In the first section we report on the agent, task, optimization technique, and analysis method. The section that follows presents an informational analysis for the best and top performing agents. In the last section we summarize the results.

\section{Methods}
%This section describes the design of the agent, task and the evolutionary algorithms and the underlying equations that govern them.

\begin{figure}
\centering
\includegraphics[height=3.5cm]{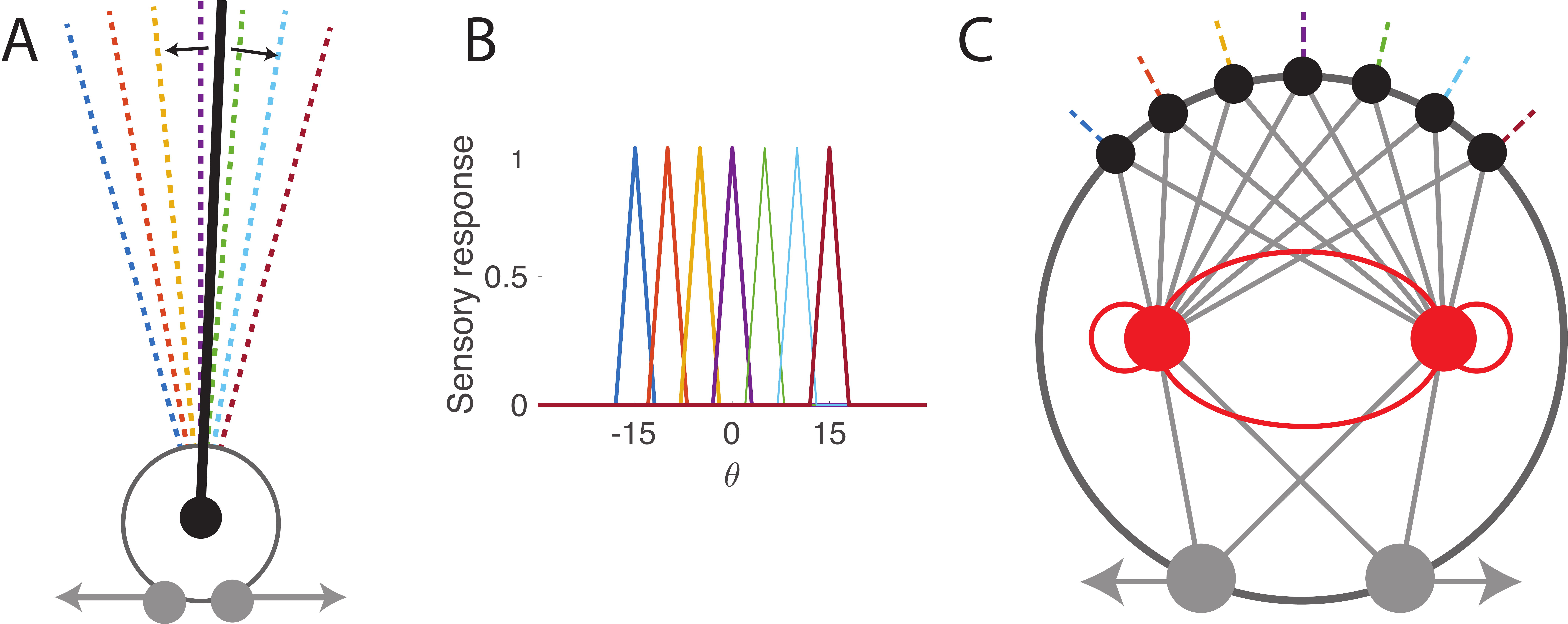}
\caption{Task set up and agent design. [A]~The agent (gray circle) moves left/right along the horizontal axis (gray arrows) sensing the pole (black rod) through the seven vision rays (color dashed lines), with a range of $36$ degrees. [B]~Sensory rays have a linearly diffused receptive field from their centers and overlap at the edges with the receptive fields of adjacent rays. [C]~The agent has $7$ vision sensors (black) connected to two recurrent spiking interneurons (red) connected to two motor units (gray).}
\label{fig:setUp}
\end{figure}

{\it Agent design.}
The agent lives in a 1-dimensional world with the pole attached to its center. Seven equidistant rays of ``vision'' with overlapping receptive fields spanning 36\degree provide it with sensory stimuli (Fig.~\ref{fig:setUp}A,B). The control network of the agent has three primary components: sensory units, spiking interneurons, and  motor units. There is one sensory units per ray, which merely pass on the signal received from the rays. Sensory units are fully connected to $N$ interneurons (here $N=2$), modeled by Izhikevich spiking neuron model \cite{Izhikevich:2003}. The model has 4 parameters per neuron and is governed by a two-dimensional system of ordinary differential equations~\cite{Izhikevich:2003}. Interneurons are recurrently connected (Fig.~\ref{fig:setUp}C). Therefore, each interneuron receives weighted input from each sensory unit, $S_i$, and from other spiking interneurons, $I_i$, as follows:

\begin{equation}
S_i+I_i = \sum_{j=1}^7 w_{ji}^s s_j + \sum_{j=1}^N w_{ji}^i o_i
\end{equation}

%\noindent and from other spiking interneurons:
%\begin{equation}
%I_i = \sum_{j=1}^N w_{ji}^i o_i
%\end{equation}

\noindent where $s_j$ is the input at the $j^{\text{\tiny th}}$ sensory unit, $w_{ji}^s$ is the strength of the connection from the $j^{\text{\tiny th}}$ sensory unit to the $i^{\text{\tiny th}}$ spiking interneuron, $w_{ji}^i$ is the strength of the recurrent connections from the $j^{\text{\tiny th}}$ to the $i^{\text{\tiny th}}$ spiking neuron, and $o_i$ is the output of the neuron. The sign of all outgoing connections from an interneuron depends on its excitatory or inhibitory nature, as identified by a binary parameter.  Finally, the layer of interneurons feeds into the two motor neurons, that has the following state equation:

\begin{equation}
\tau_m \dot{m_i} = - m_i + \sum_{j=1}^N w_{ji}^m \bar{o}_j \quad
i = 1,2
\end{equation}

\begin{equation}
\bar{o}_j(t)= \frac{1}{h_j} \sum_{k=0}^{h_j} o_j(t-k)
\end{equation}

\noindent where $m_i$ represents the motor neuron state, $w_{ji}^m$ is the strength of the connection from the $j^{\text{\tiny th}}$ spiking interneuron to the $i^{\text{\tiny th}}$ motor neuron, $\bar{o}_j$ represents the firing rate code, the moving average over a window of length $h_j$ for the output of spiking interneuron $j$. Finally, the difference in output between the motor neurons results in a net force that acts upon the agent, moving it along the track. The network was simulated using Euler integration with step size 0.01.

{\it Pole Balancing Task Design.}
The agent can move left/right to balance a pole for as long as possible. The pole-balancing task was implemented based on published descriptions~\cite{Barto:1983}. The force from the agent, computed as the difference between motor unit outputs, affects the angular acceleration of the pole and acceleration of the agent. The physical parameters such as mass, length and coefficient of friction were all set as per the same reference. While typically pole-balancers receive as input the angle of the pole ($\theta$), its angular velocity ($\omega$), the agent's position ($x$) and velocity ($v$), our implementation was modified to only sense the pole through the sensory rays.

%set as follows - $m=.1Kg$ is the mass of the pole, $m_a=1Kg$ is the mass of the agent, $F$ is the force, $l=0.5m$ is the half-length of the pole, $\mu_c=0.0005$ is the coefficient of friction between the agent and the track, $\mu_p=0.000002$ is the coefficient of friction between the agent and the pole, $\theta$ is the angle of the pole, $\omega$ the angular velocity, $g$ is the acceleration due to gravity $\dot{v}$ is the agent acceleration, $v$ its velocity, $x$ is the agent position and $sgn(v)$ is the signum function.

%\begin{equation}
 %   \dot{\omega} = \frac{gsin\theta + \frac{(\mu_csgn(v) - F - ml\omega^2sin\theta)cos\theta}{m+m_a} + \frac{\mu_pw}{ml}}{l(\frac{4}{3} - \frac{m}{m+m_a}cos^2\theta)}
%\end{equation}

%\begin{equation}
%    \dot{v} = \frac{F + ml(w^2sin\theta - \dot{w}cos\theta) - \mu_csgn(v)}{m+m_c}

%\end{equation}

{\it Evolutionary Algorithm.}
The network was optimized using a real valued evolutionary algorithm. A solution was encoded by 38 parameters, including the intrinsic parameters of the Izhikevich neurons, the neuron-specific size of the window for estimating rate code, all connection weights (sensory-interneuron, interneuron-interneuron, interneuron-motor), and the time constant, bias and gain for the motor units. Parameters were constrained to certain ranges: connection strengths $\in$ $[-50,50]$; motor unit biases $\in$  $[-4,4]$; time-constants $\in$  $[1,2]$. The range of intrinsic parameters and the polarity of the outgoing weights from the inter-neuron depended on a binary inhibitory/excitatory neuron flag parameter in the genotype~\cite{Izhikevich:2003}. The population consisted of 100 individuals.
%and reproduction was carried out by selecting parents based on a $0.04$ elitist fraction followed by a uniform crossover (probability $0.5$) and a Gaussian mutation variance of $\mu=0$,$\sigma^2=0.5$.

{\it Fitness Function.}
Performance was estimated by averaging over 16 trials, starting at pole angles $\theta_0$ between $\pm12\degree$, in increments of 3\degree, and two initial angular velocities, $\omega_0=\pm0.001$. The fitness function to be maximized was $f=(\sum_{t=1}^{T} cos(\theta_t))/T$, where $T=500s$ is the maximum duration of the run. The pole was considered balanced if it was within the sensory range of the agent. Also, the track length was fixed at $45$ units, beyond which the trial was terminated.

{\it Mutual Information.}
The amount of information contained in one variable about another was estimated using Mutual Information (MI). We quantified the information that neurons contain about pole angle ($\theta$), angular velocity ($\omega$), agent position ($x$) and agent velocity ($v$) by calculating their probability distributions (using a binning method with each bin of width 0.01):

\begin{equation}
    MI(N,X) = \sum_{n \in N}\sum_{x \in X} p(x,n)log\frac{p(x,n)}{p(x) p(n)}
\end{equation}

\section{Results}
%This section presents the results of the evolutionary optimization. We focus on an information theoretic analysis of the best evolved agent first and on the top 10 evolved agents second.

\begin{figure}
\centering
\includegraphics[height=3.5cm]{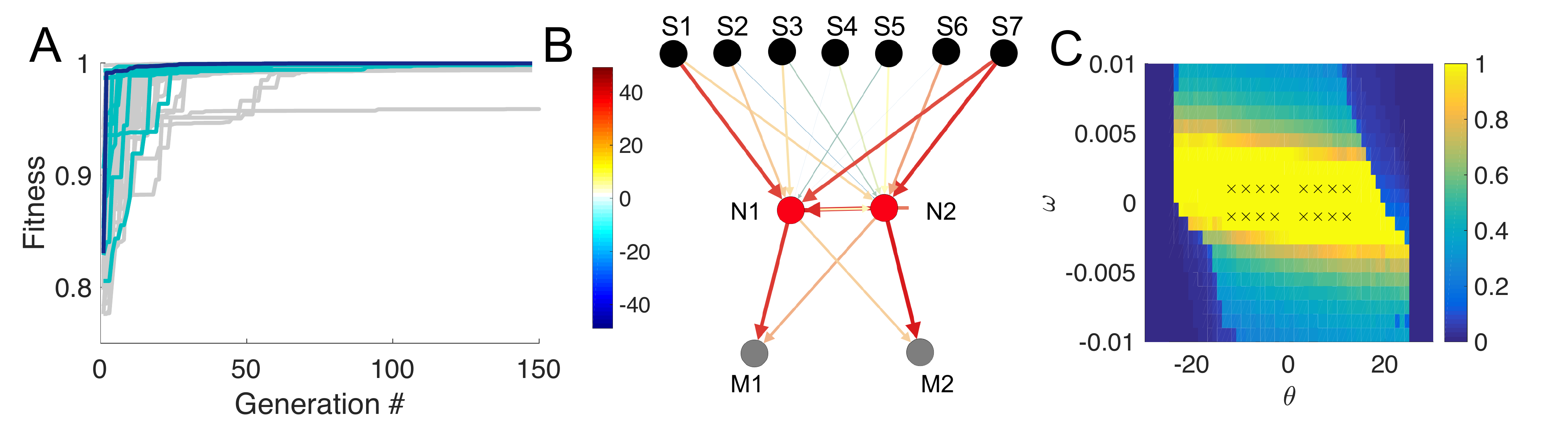}
\caption{Optimization performance. [A]~Fitness of the best individual in the population vs. generations for 100 evolutionary runs. Best run in blue, top 10 in light blue and the rest in gray. [B]~Network structure of the best agent. The width of the edges indicate the magnitude of the weights and are also color coded for polarity. The sensory units in black are identified as S1-S7, spiking interneurons N1,N2 in red and motor units M1,M2 in grey. [C]~Generalization performance over a broader and finely grained set of initial conditions that were tested during evolution (marked by cross marks).}
\label{fig:bestAgent}
\end{figure}

{\it Performance of evolutionary optimization.}
While pole balancing is a well-known benchmark, it was also a relatively easy task to optimize. The evolutionary algorithm found fairly good solutions (around 75\% performance) at the very first random draw of the population. Fig.~\ref{fig:bestAgent}A shows the performance of the best agent in every generation over time. Out of the $100$ evolutionary runs, $99$ converged to over $99\%$ fitness with only two spiking interneurons.

\begin{figure}
    \centering
    \includegraphics[height=11.5cm]{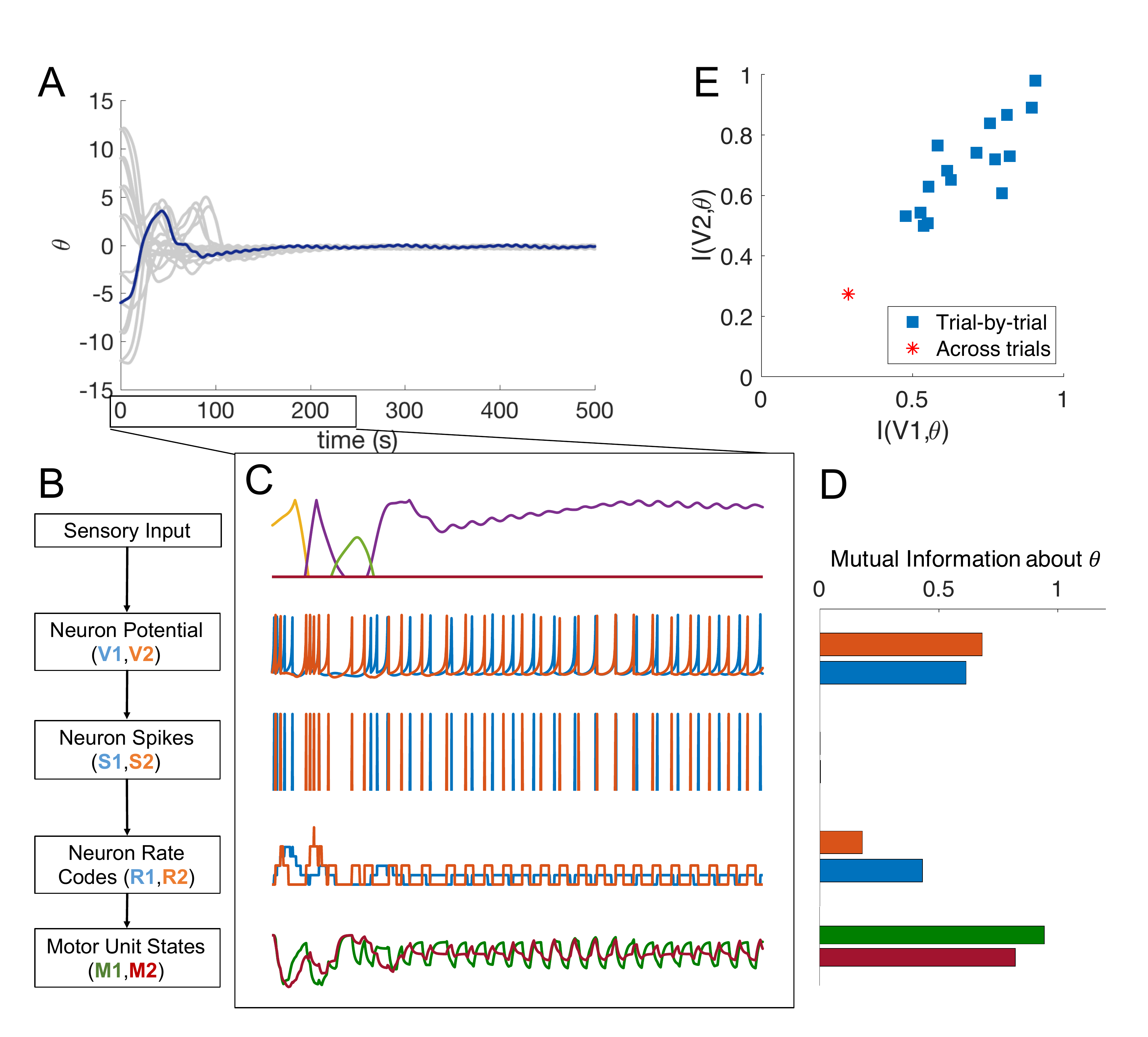}
    \caption{Behavior of the best agent and one of its trials in detail. [A]~Angle of the pole over time on 16 trials. One of the trials ($\theta_0=-6\degree$ and $\omega_0=0.001$) is highlighted and explored further.  [B]~Information flowchart: from sensory input rays through the spiking inter-neuron layer, composed of 3 levels of processing (neuron potential, spiking activity and rate code), and then the motor units. [C]~A sample trace for each of the components corresponding to each box in B from the trial highlighted in A. Each color in the first figure is matched with the sensory rays in Fig.~\ref{fig:setUp}. The blue and orange colors indicate interneurons 1 and 2 respectively. The green and brown traces corresponds to the left and right motor neurons respectively. [D]~Mutual Information about the pole angle ($\theta$) for the highlighted trial for each of the components. [E]~Mutual information about $\theta$ in the neuron potential ($V$). A comparison between the trial-by-trial MI vs. the total MI across trials.}
    \label{fig:bestInfoArch}
\end{figure}

\begin{figure}
    \centering
    \includegraphics[height=3.2cm]{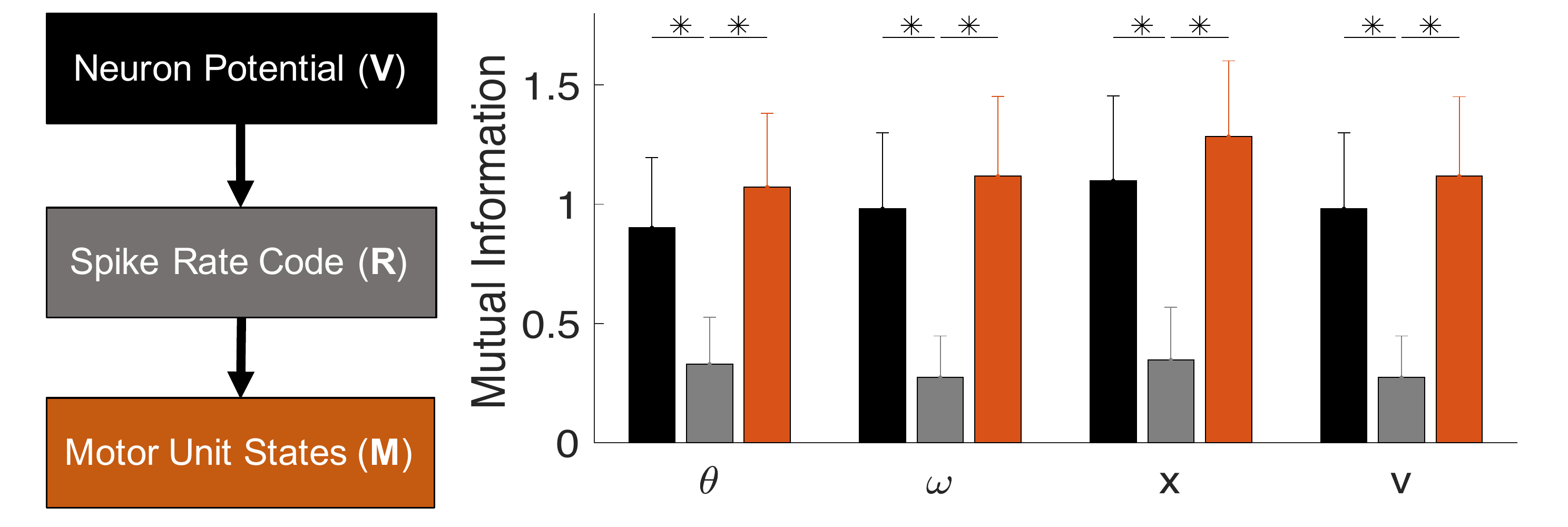}
    \caption{MI about the four environmental variables: pole angle ($\theta$), pole angular velocity($\omega$), agent position ($x$) and agent velocity ($v$) in the network elements: neuron potentials ($V$), rate coded outputs of neurons ($R$), and motor unit states ($M$). Paired samples t-test yielded highly significant ($p\ll0.05$) differences between the information contained about $\theta, \omega$ and $v$ in $V$ versus $R$ and $R$ versus $M$.}
    \label{fig:vrSignificance}
\end{figure}

{\it Network structure and performance of the best agent.}
The network structure of the best agent from the $100$ runs is shown in Fig.~\ref{fig:bestAgent}B.
%Interestingly, the agent evolved to be close to bilaterally symmetric and had an architecture with semblance to the Braitenberg's vehicle alluding to the reactive nature of this task.
The behavioral traces of this agent on the 16 trials specified in the previous section are shown in Fig.~\ref{fig:bestInfoArch}A. This agent achieved a fitness of $99.4\%$. To test for generalization, its performance was evaluated on a finer and broader range of conditions, post-optimization (Fig.~\ref{fig:bestAgent}C): initial pole angle $\theta_0$ in the range $[-45,45]$ and initial pole angular velocity $\omega_0$ in the range $[-0.01,0.01]$. As can be seen, the agent generalizes well within and outside the region it was evolved on. Note that $\theta_0$ that are beyond $18\degree$ on either side are beyond the range of the sensory rays. %Generalization was computed as the normalized sum of all these fitness i.e. A fitness of 1 all over the map would give a generalization value of 1. This agent has a generalization value of 0.45.

{\it Encoding of environmental variables.}
The different network elements that manipulate the input are shown in Fig.~\ref{fig:bestInfoArch}B. Sensory signals first act on neurons'  potential. The neuron fires based on dynamics in its potential, which is then interpreted by its rate. The motor units then convert this discrete spike rate to smooth continuous movement. Note that a single neuron has three levels of informational content - continuous valued potential, binary spikes, and discrete spike rate codes. Fig.~\ref{fig:bestInfoArch}C shows traces for the highlighted trial in Fig.~\ref{fig:bestInfoArch}A of the best agent. Although $\theta$ is the most directly available information, unlike the standard practice of directly providing it, the sensory rays provide an agent-centric perspective of $\theta$. MI between each of the network elements with $\theta$, see Fig.~\ref{fig:bestInfoArch}D, revealed that internal potential of neurons have relatively more information about $\theta$ than the spike rate and so does the motor units. Albeit only for one trial in one agent, this shows that the bottleneck does not always become narrow in control tasks. MI also reveals that indirect encoding of $\omega$, $x$ and $v$ all happen in the very first stage of the network, neuron potential (black bars in Fig.~\ref{fig:vrSignificance}). This can be attributed to the recurrent nature of connections between the interneurons and also their rather complex non-linear internal dynamics.

{\it Analysis of the Information Bottleneck.}
All available information, as shown by MI in $V$, is not necessarily used in controlling movement, as shown by relatively lower information in $R$. To further study the bottleneck, we compared the amount of information contained in neuron potentials, $V$, versus the rate coded outputs of the neurons, $R$. Note that the spikes themselves do not have any information about the environment but, in fact, encode them in its rate. For each of the environmental variables a paired samples t-test was conducted with a significance threshold of $p<0.05$. This revealed that there is a significant drop in the amount of information between $V$ and $R$ (Fig.~\ref{fig:vrSignificance}) robustly across the top ten agents. This can be attributed to  the loss due to the discretization of information available in $V$ as spikes. However, the information in $R$ is sufficient to perform the behavior with great accuracy and so this is in fact an efficient encoding of information. The minimal yet relevant information that is encoded in $R$ is interpreted by the motor units. They integrate $R$ from the interneurons and their outputs directly impact $\theta$, $\omega$, $x$ and $v$ and so this is where a deviation from the IB principle is expected. Statistical analyses of the MI between $R$ and the motor units state, $M$, using the paired samples t-test yielded highly significant ($p\ll0.05$) increase in information about all environmental variables in $M$ (Fig.~\ref{fig:vrSignificance}). This shows that the IB for control tasks is not always a filtering of information but is rather filtering followed by an expansion at the control layer.

{\it Context sensitive information encoding.}
From previous analysis, we know that components of the network encode information about the environment. But what information do they encode? Typically, when a neuron is said to encode information about a feature of the environment it is thought to be a consistent, context-independent code. To explore this idea further, we compared the MI the neuron potential has about the pole's angle $I(V,\theta)$ on a trial by trial basis to the information that same neuron has across all trials about the same feature of the environment on the best agent (Fig.~\ref{fig:bestInfoArch}E) and the top 10 agents. A one-sample t-test of the combined MI with the distribution of trial-by-trial MI values yielded a highly significant difference ($p\ll0.05$). This means that the combined information is significantly lower than trial-by-trial information, and therefore that encoding is highly context-dependent across all evolved pole-balancers.

\section{Discussion}
In this paper, we have presented results from an information theoretic analysis of recurrent spiking neural networks that were evolved to perform a continuous control task: agent-centric non-Markovian pole balancing. Our results can be summarized as follows. First, networks with as few as two spiking neurons could be evolved to perform this task. Second, through the use of MI, we show that the spiking network encoded environmental variables of interest that were directly and indirectly available. Third, we show that the information bottleneck from the neuron potential to its firing rate is an efficient filtering/compression, which was followed by an increase in information at the control layer on account of their causal effect on the environment. This is a phenomenon that we expect to arise in control tasks in general, and plan to explore further with different tasks and types of networks. Perhaps, this can develop into an optimization method for neural network control. Finally, we show that the information encoded in the spiking neurons vary across trials, causing the across-trial combined information to be significantly lower. This can mean either that the same stimuli are encoded in different ways (redundancy) or that different stimuli are mapped on to the same encoding (generalization) or both. This warrants further analysis to understand the encoding in more detail and more interestingly, to understand how the context helps disambiguate the generalized representations during a trial.


\begin{thebibliography}{4}

\bibitem{Bengio:2013}Bengio, Y., Courville, A., \& Vincent, P. (2013). Representation learning: A review and new perspectives. IEEE transactions on pattern analysis and machine intelligence, 35(8), 1798-1828.

\bibitem{Krizhevsky:2012}Krizhevsky, A., Sutskever, I., \& Hinton, G. E. (2012). Imagenet classification with deep convolutional neural networks. In Advances in neural information processing systems (pp. 1097-1105).

\bibitem{Mnih:2013}Mnih, V., Kavukcuoglu, K., Silver, D., Graves, A., Antonoglou, I., Wierstra, D., \& Riedmiller, M. (2013). Playing atari with deep reinforcement learning. arXiv preprint arXiv:1312.5602.

\bibitem{Silver:2016}Silver, D., Huang, A., Maddison, C. J., Guez, A., Sifre, L., Van Den Driessche, G., ... \& Dieleman, S. (2016). Mastering the game of Go with deep neural networks and tree search. Nature, 529(7587), 484-489.

\bibitem{Tishby:2000}Tishby, N., Pereira, F. C., \& Bialek, W. (2000). The information bottleneck method. arXiv preprint physics/0004057.

\bibitem{Berger:1971}Berger, T. (1971). Rate‐Distortion Theory. Encyclopedia of Telecommunications.

\bibitem{Tishby:2015}Tishby, N., \& Zaslavsky, N. (2015, April). Deep learning and the information bottleneck principle. In Information Theory Workshop (ITW), 2015 IEEE (pp. 1-5).

\bibitem{Simoncelli:2001}Simoncelli, E. P., \& Olshausen, B. A. (2001). Natural image statistics and neural representation. Annual review of neuroscience, 24(1), 1193-1216.

\bibitem{Borst:1999}Borst, A., \& Theunissen, F. E. (1999). Information theory and neural coding. Nature neuroscience, 2(11), 947-957.

\bibitem{Pasemann:1997}Pasemann, F., \& Dieckmann, U. (1997). Evolved neurocontrollers for pole-balancing. Biological and Artificial Computation: From Neuroscience to Technology, 1279-1287.

\bibitem{Onat:1998}Onat, A., Nishikawa, Y., \& Kita, H. (1998). Q-learning with recurrent neural networks as a controller for the inverted pendulum problem.

\bibitem{Barto:1983}Barto, A. G., Sutton, R. S., \& Anderson, C. W. (1984).Neuron-like adaptive elements that can solve difficult learning control problems. Behavioural Processes.9,89.

\bibitem{Izhikevich:2003}Izhikevich, E. M. (2003). Simple model of spiking neurons. IEEE Transactions on neural networks, 14(6), 1569-1572.

\bibitem{Miikkulainen:2017} Miikkulainen, R., Liang, J., Meyerson, E., Rawal, A., Fink, D., Francon, O., ... \& Hodjat, B. (2017). Evolving Deep Neural Networks. arXiv preprint arXiv:1703.00548.

\bibitem{Salimans:2017}Salimans, T., Ho, J., Chen, X., \& Sutskever, I. (2017). Evolution Strategies as a Scalable Alternative to Reinforcement Learning. arXiv preprint arXiv:1703.03864.

\bibitem{Jong:2006}De Jong, K. A. (2006). Evolutionary computation: a unified approach. MIT press.
\end{thebibliography}
\end{document}